\documentclass{article}

\usepackage{microtype}
\usepackage{graphicx}
\usepackage{subfigure}
\usepackage{booktabs} 

\usepackage{multirow}
\usepackage{makecell}
\usepackage{tablefootnote}
\usepackage[para,online,flushleft]{threeparttable}

\usepackage{hyperref}

\usepackage{CJKutf8}

\usepackage[accepted]{icml2024}

\usepackage{amsmath}
\usepackage{amssymb}
\usepackage{mathtools}
\usepackage{amsthm}

\usepackage[capitalize,noabbrev]{cleveref}

\theoremstyle{plain}

\theoremstyle{definition}

\theoremstyle{remark}

\usepackage[textsize=tiny]{todonotes}

\icmltitlerunning{Is Self-knowledge and Action Consistent or Not:         Investigating Large Language Model's Personality}

\begin{document}

\twocolumn[
\icmltitle{Is Self-knowledge and Action Consistent or Not:  \\
           Investigating Large Language Model's Personality}




\begin{icmlauthorlist}
\icmlauthor{Yiming Ai}{yyy}
\icmlauthor{Zhiwei He}{yyy}
\icmlauthor{Ziyin Zhang}{yyy}
\icmlauthor{Wenhong Zhu}{yyy}
\icmlauthor{Hongkun Hao}{yyy}
\icmlauthor{Kai Yu}{zzz}
\icmlauthor{Lingjun Chen}{xxx}
\icmlauthor{Rui Wang}{yyy}
\end{icmlauthorlist}

\icmlaffiliation{yyy}{MT Lab, Department of Computer Science and Engineering, Shanghai Jiao Tong University, Shanghai, China}
\icmlaffiliation{xxx}{School of Education, Shanghai Jiao Tong University, Shanghai, China}
\icmlaffiliation{zzz}{X-LANCE Lab, Department of Computer Science and Engineering, Shanghai Jiao Tong University, Shanghai, China}

\icmlcorrespondingauthor{Rui Wang}{wangrui12@sjtu.edu.cn}

\icmlkeywords{Psycholinguistics, Self-knowledge-Action Congruence, Benchmark, Personality Traits}

\vskip 0.3in
]



\printAffiliationsAndNotice{}  

\begin{abstract}
In this study, we delve into the validity of conventional personality questionnaires in capturing the human-like personality traits of Large Language Models (LLMs). Our objective is to assess the congruence between the personality traits LLMs claim to possess and their demonstrated tendencies in real-world scenarios. By conducting an extensive examination of LLM outputs against observed human response patterns, we aim to understand the disjunction between self-knowledge and action in LLMs.
\end{abstract}

\begin{CJK*}{UTF8}{gbsn}
\section{Introduction}

Personality, a foundational social, behavioral phenomenon in psychology, encompasses the unique patterns of thoughts, emotions, and behaviors of an entity \citep{allport1937personality, roberts2022personality}. In humans, personality is shaped by biological and social factors, fundamentally influencing daily interactions and preferences \citep{roberts2007power}. Studies have indicated how personality information is richly encoded within human language \citep{goldberg1981language, saucier2001lexical}. LLMs, containing extensive socio-political, economic, and behavioral data, can generate language that expresses personality content. Measuring and verifying the ability of LLMs to synthesize personality brings hope for the safety, responsibility, and coordination of LLM efforts \citep{gabriel2020artificial} and sheds light on enhancing LLM performance in specific tasks through targeted adjustments.

Thus, evaluating the anthropomorphic personality performance of LLMs has become a shared interest across fields such as artificial intelligence(AI) studies, social sciences, cognitive psychology, and psychometrics. A common method for assessment involves having LLMs answer personality questionnaires \citep{huang2024chatgpt}. However, the reliability of LLMs' responses, whether the responses truly reflect LLMs' genuine personality inclinations, and whether LLMs' behavior in real-world scenarios aligns with their stated human-like personality tendencies remain unknown.

To illustrate such inconsistency in LLMs, we introduce two concepts: \textit{self-knowledge} \footnote{\url{https://plato.stanford.edu/entries/self-knowledge/}} and \textit{action}. In the following, \textit{self-knowledge} specifically refers to an individual's understanding and awareness of their own internal states, including personality, emotions, values, motivations, and behavioral patterns. The term \textit{personality knowledge} mentioned later is equivalent to self-knowledge. \textit{Action} refers to the behavioral state of an individual in actual situations. For humans, action is the way self-knowledge is transformed into external expression. Self-knowledge and action are meant to be two interacting aspects.

From the perspective of LLMs, a discordance between an LLM's asserted self-knowledge and its action can result in noteworthy adverse outcomes. For example, while an LLM may claim to prioritize human friendliness, its failure to manifest amicable behaviors in real-world situations is undoubtedly a circumstance we fervently seek to avert. Hence, our study endeavors to assess the alignment between the personality traits claimed by LLMs and their actual behavior tendency. From the perspective of personality scales, there have been several studies investigating the reliability of personality questionnaires on LLMs \citep{huang2023revisiting, safdari2023personality}. However, there has yet to be any exploration of the validity of psychological scales on LLMs. Our work aims to address this gap in the research literature.
In general, our research makes three significant contributions:

\begin{itemize}
    \item We design a behavior tendency questionnaire that reflects real-world situations and behaviors based on them; 
    \item We evaluate the self-knowledge-action congruence of LLMs, revealing substantial disparities between LLMs' personality knowledge and behavioral inclinations;
    \item We empirically test various LLMs against observed human response patterns and formulate conjectures, thereby shedding light on the potential and limitations of LLMs in mimicking complex human psychological traits.
\end{itemize}


In Section \ref{sec-cd}, we introduce the process of our corpus design. Section \ref{sec k-a} presents the our empirical analysis – evaluating self-knowledge-action congruence of various LLMs. Finally, in Section \ref{sec-con}, we conclude our work. 

\section{Corpus Design}\label{sec-cd}

In the nuanced exploration of anthropomorphic personality traits within LLMs, selecting the most appropriate personality tests is paramount. Among diverse personality assessments, the comprehensive coverage of personality dimensions, theoretical robustness, and practical relevance make the Big Five Personality Traits \citep{goldberg1981language, costa2008revised} and the Myers-Briggs Type Indicator (MBTI) \citep{myers1962myers} the most fitting choices for our study. 

To devise a straightforward yet impactful evaluation of LLMs' personality traits, we've opted for two questionnaires (TDA-100 \citep{goldberg1992development} and BFI-44 \citep{john1991big}) rooted in the Big Five model, along with one questionnaire (16 Personalities \footnote{\url{https://www.16personalities.com/}}) based on the MBTI model. These selections were made due to their proven high reliability and validity in both English and Chinese \citep{goldberg1992development, john1991big, makwana2020confirmatory, zhang2012BFC}. Based on these questionnaires, we ensure that our investigation into the anthropomorphic traits of LLMs is grounded in robust psychological methodology and thereby construct a bilingual personality knowledge questionnaire, including a total of 180 statements.


In the following, we will detail the methodology adopted to create a comprehensive corpus aimed at evaluating the congruence between the personality traits professed by LLMs and their behavior tendency. The corpus is comprised of 2 parts: a personality knowledge questionnaire and a behavior tendency questionnaire. The former includes 180 statements mentioned before, and the latter is closely aligned with the former.

We apply the common method of constructing behavioral procedures approach test, \textbf{sample approach}, which assumes that the test behavior constitutes a subset of the actual behaviors of interest \citep{golfried1972traditional}. The detailed design process is outlined as follows:

\textbf{Step 1:} As \citet{golfried1972traditional} has mentioned that the ideal approach to response expression would constitute the individual's actual response in a real-life situation, in that this represents the most direct approach to behavioral sampling. We recruited 16 individuals, each representing a distinct MBTI type, to undertake the following task: for every statement in the personality knowledge questionnaire, they provided a \textit{practical scenario case}. Each scenario case comprises situations drawn from their own lives, along with two completely contrasting actions: Action A and Action B. Action A fully aligns with the statement, while Action B completely contradicts it. The content of Action A and Action B need to be kept basically the same length.

\textbf{Step 2:} Following the acquisition of the 16 practical scenario cases corresponding to each statement, we condensed them into a single case. For 19 statements exhibiting significant variations in cases, we amalgamated them into 2 to 3 cases. 

\textbf{Step 3:} For statements associated with multiple practical scenario cases, we tasked the previously enlisted 16 individuals to assign ratings to each case. A rating of 1 was given if they believed the case accurately reflected the meaning of the corresponding statement in the personality knowledge questionnaire; otherwise, a rating of 0 was assigned. The case with the highest score for these 19 statements was selected as the final practical scenario case.

\textbf{Step 4:} We enlisted the participation of 10 reviewers to assess the consistency of the 180 \textit{personality knowledge - practical scenario} pairs. The results demonstrate that the consistency approval rate for each pair exceeds 90\%. 

All the individuals involved are native Chinese speakers with a level of English proficiency of CEFR C1. The detailed instructions for the scenario providers and reviewers are shown in Appendix \ref{sec:human}. Several examples of a \textit{personality knowledge - practical scenario} pair are shown in Appendix \ref{ap-es}.

The culmination of this meticulous process is a bilingual English-Chinese Parallel Sentence Pair Self-knowledge-Action Test Set, comprising 180 matched pairs of personality knowledge and action scenarios. This corpus serves as a fundamental tool in our study, allowing us to rigorously evaluate the LLMs' proficiency in understanding and acting upon various personality traits, bridging the gap between personality understanding and practical action in the realm of AI.

\section{Experiment on LLMs' Self-knowledge-Action Congruence} \label{sec k-a}

\subsection{Experiment}

Among all LLMs, we selected baize-v2-7b, ChatGLM3, GPT-3.5-turbo, GPT-4, internLM-chat-7b, Mistral-7b, MPT-7b-chat, Qwen-14b-chat, TULU-2-DPO-7b, Vicuna-13b, Vicuna-33b and Zephyr-7b, 12 LLMs in total, who could answer the personality cognitive questionnaire in the form of a Q$\&$A. The detailed setup is shown in Appendix \ref{ap-es}. Then, we rewrote a prompt for LLM to answer the former part of our corpus - personality knowledge questionnaire based on the response requirements of the MBTI-M questionnaire \citep{gu2012mbti} in Appendix \ref{ap-es}.

Upon reviewing the responses from the LLMs, we discovered that some LLMs failed to grasp the intended meaning of the prompts, resulting in unreasonable responses as detailed in Appendix \ref{ap-ur}. Out of the LLMs assessed, only seven LLMs, ChatGLM3, GPT-4, GPT-3.5-turbo, Mistral-7b, Vicuna-13b, Vicuna-33b, and Zephyr7b, produced valid responses. Subsequently, we sifted through these valid responses, computed their averages to represent the LLMs' actual responses, and proceeded to evaluate the reliability of these responses, as outlined in Appendix \ref{sec-reliability}. Following this assessment, we determined that the responses from \textbf{ChatGLM3, GPT-3.5-turbo, GPT4, Vicuna13b} and \textbf{Vicuna33b} are reliable for further personality analysis.

In the following, we explore the alignment between responses given by LLMs to personality knowledge questionnaires and their actions within designed scenarios. Regarding the prompt for questioning, we selected the instructions of five common academic questionnaires with effective analysis of reliability and validity \citep{makwana2020confirmatory, johnson1998third, goldberg1992development, john1991big, nardi2011neuroscience}, 16 Personalities Test, MBTI-M, TDA-100, BFI-44-Children adapted and Dario Nardi's Cognitive Test, as the prompt for the LLM of questioning of the personality knowledge questionnaire. We utilize various prompts to prevent any particular prompt from exerting a specific influence on LLM responses, thereby accurately reflecting the general tendencies of LLMs when answering personality knowledge questionnaires.


As for the responding approach to the personality knowledge questionnaire, according to the structure of the chosen personality scales in \ref{sec-cd}, responses to statements are initially mapped on a 7-point Likert scale, ranging from 1 to 7.  According to several previous studies, when responding to personality scales, LLMs' answers often remain consistent, regardless of factors such as question order, quantity, answer sequence, or timing of inquiry. \citep{huang2023revisiting, safdari2023personality}. Therefore, for each prompt, we asked each LLM 10 times with the original form of our chosen personality scales and then screened the valid responses. We averaged all the valid responses to reduce errors and reflect the general LLMs’ response pattern. and rounded the average response to each statement to the nearest whole number as each LLM's response to the personality knowledge questionnaire. The details of the prompts are shown in Appendix \ref{ap-es}.

Concerning the prompt for LLM to answer the latter part of our corpus-behavior tendency questionnaire, we inherit the instruction of the MBTI-M questionnaire \citep{gu2012mbti} and rewrite it, for we intend to change the responding approach.

We apply a 7-point graded forced-choice format \citep{brown2018ordinal} as the responding approach. Currently, the commonly used response formats for questionnaires in psychometrics are the forced-choice format \citep{sisson1948forced} and the Likert scale format \citep{joshi2015likert}. In comparison to traditional forced-choice scales, graded forced-choice scales exhibit comparable validity, superior reliability and model fit. Contrary to Likert scales, graded forced-choice scales show better model fit and slightly higher self-other agreement \citep{zhang2023moving}. The specific meaning of numbers in common 7-point graded forced-choice is shown in Appendix \ref{ap-mfc}.

Here, given that we have rewritten the prompt of responding to personality knowledge questionnaire based on the original instructions of the chosen personality scales, thereby not indicating the specific meaning of numbers 2, 3, 5 and 6. We followed this prompt pattern to avoid influence on LLMs' responses brought by such change, which means only retain the meaning of numbers 1, 4 and 7. Hence, the specific prompt is: \textit{Read the following scenarios with actions A and B carefully and rate each scenario in the range from 1 to 7. 1 means that action A applies to you completely in this scenario, 4 means that action A and action B equally apply (or not) to you in this scenario, and 7 means that action B applies to you completely in this scenario. You only need to give the number. }

These measures above allow us to to observe the congruence between self-knowledge and action of LLMs, to compare human and LLM responses.

\begin{table*}[h]
\caption{\label{table-task2} LLMs' Self-knowledge - Action Congruence Performance with Reference of Human Respondents' Performance (AVG, SD, MIN and MAX represents the average number, standard deviation, minimum and maximum.)}
\vskip 0.15in
\centering
\begin{tabular}{lcccc}
\toprule
\multirowcell{2}{\bf LLMs $\&$ Human\\\bf Respondents} & \multirowcell{2}{\bf Cosine \\\bf Similarity} & \multirowcell{2}{\bf Spearman Rank\\\bf Correlation Coefficient} & \multirowcell{2}{\bf Value Mean\\\bf Difference} & \multirowcell{2}{\bf Proportion of\\\bf Consistent Pairs} \\
             &  &  &  &   \\
\midrule
ChatGLM3    & 0.24 & 0.23 & 1.58 & 47.22\%  \\
GPT-3.5-turbo    & 0.17 & 0.19 & 1.74 & 50.56\%  \\
GPT-4            & 0.52 & 0.56 & 1.02 & 78.89\%  \\
Vicuna-13b        & 0.08 & 0.07 & 1.57 & 52.78\%  \\
Vicuna-33b        & 0.18 & 0.06 & 1.68 & 52.22\%  \\
LLMs(AVG $\pm$ SD)         & \bf 0.24 $\pm$ 0.15 & \bf 0.22 $\pm$ 0.18 & \bf 1.52 $\pm$ 0.26 & \bf 56.78 $\pm$ 11.25\%  \\
\midrule
Human(AVG $\pm$ SD)             & \bf 0.76 $\pm$ 0.09 & \bf 0.78 $\pm$ 0.08 & \bf 0.69 $\pm$ 0.27 & \bf 84.69 $\pm$ 8.22\%  \\
Human(MIN)            & 0.61 & 0.66 & 1.08 & 73.78\%   \\
Human(MAX)            & 0.95 & 0.96 & 0.07 & 99.44\%  \\
\bottomrule
\end{tabular}
\vskip -0.1in
\end{table*}

\subsection{Results}\label{results}
To quantify the similarity between responses, we employ the following four metrics: cosine similarity, Spearman's rank correlation coefficient, value mean difference (VMD) and Proportion of Consistent Pairs.

\textbf{Cosine Similarity} \quad A measure used to calculate the cosine of the angle between two vectors in a multi-dimensional space, offering a value range from -1 (exactly opposite) to 1 (exactly the same), where higher values indicate greater similarity.
\begin{equation}
    s_{\rm{cos}} = 
\begin{array}{cl}
\sum_{i=1}^n\left(x_i\times y_i\right)\\
\hline
\sqrt{\sum_{i=1}^n\left(x_i\right)^2} \times \sqrt{\sum_{i=1}^n\left(y_i\right)^2}
\end{array} ,
\end{equation}%
where $x_i$ are LLMs' responses of personality knowledge questionnaire, $y_i$ are LLMs' corresponding responses of scenario and action questionnaire, and $x_i$ and $y_i$ correspond to each other one-to-one.

\textbf{Spearman's Rank Correlation Coefficient} \quad A non-parametric measure of rank correlation, assessing how well the relationship between two variables can be described using a monotonic function. Its value ranges from -1 to 1, where 1 means a perfect association of ranks. Specifically, we rank the responses on two questionnaires of the LLMs based on their numerical values separately. Then, we calculate the difference in rankings for each personality knowledge – scenario $\&$ action pair. Afterwards, we use the following formula to calculate the coefficient $r_s$. 
\begin{equation}
    r_s=1-\frac{6\sum d_i^2}{n(n^2-1)},
\end{equation}
where $d_i$ is the difference in rankings of each pair and $n$ is the total count of pairs.

\textbf{Value Mean Difference (VMD)} \quad Value Mean Difference is the average difference in responses across all paired items in the questionnaires, as shown in the formula below.
\begin{equation}
    \rm{VMD} = \frac{\sum d_i}{n},
\end{equation}
where $d_i$ is the difference of responses in each pair.

\textbf{Proportion of Consistent Pairs} \quad Recognizing that minor discrepancies are natural when comparing psychological tendencies with actual actions, this metric quantifies the proportion of item pairs with a response difference of 1 or less, focusing on the consistency of tendencies rather than exact matches.
\begin{equation}
    P_c = \frac{N_c}{N_t},
\end{equation}
where $N_c$ is the number of consistent pairs, $N_t$ is the total number of pairs.

For this study, we recruited 16 participants, comprising 8 males and 8 females, all native Chinese speakers with an English proficiency level of CEFR C1. As shown in Table \ref{table-task2}, the analysis of their response data yielded an average Cosine Similarity and Spearman's Rank Correlation Coefficient above 0.75, with a Value Mean Difference around 0.68, and a Proportion of Consistent Pairs exceeding $84\%$. These results indicate a high degree of similarity and strong correlation between responses to the two types of questionnaires, suggesting a basic consistency in human self-knowledge and an ability to align self-knowledge with action in real-life scenarios.

The same questionnaires were administered to the 5 LLMs selected in Section \ref{sec-reliability}, and their responses were analyzed using the aforementioned metrics. Compared to human respondents, the similarity in LLMs' responses is notably lower, and the corresponding significance test is shown in Appendix \ref{sf}. Specifically, the average Cosine Similarity and Spearman's Rank Correlation Coefficient for LLMs are substantially below those of human respondents, with a huge difference exceeding 0.42. The Value Mean Difference for LLMs averages around 1.52, indicating a substantial divergence in self-knowledge between the two types of questionnaires for LLMs. And as for most LLMs, the proportion of consistent pairs falls below 55$\%$, raising questions about LLMs' ability to achieve self-knowledge-action unity in practice.

\section{Conclusion} \label{sec-con}
We demonstrate that while LLMs exhibit some capacity to mimic human-like tendencies, there are significant gaps in the coherence between their stated personality and exhibited behaviors. This disparity probably suggests a limitation in LLMs' ability to authentically replicate human personality dynamics. Our study underscores the importance of further exploration into enhancing LLMs' ability to perform more genuinely human-like interactions, suggesting avenues for future research in improving the psychological realism of LLM outputs.

\section*{Limitations}
In this study, we delve into the alignment between what Large Language Models (LLMs) claim and their actions, aiming to discern if there's a consistency in their self-knowledge and their actual behavior tendency. This observation is merely one among several hypotheses exploring the root causes of this inconsistency, underscoring the need for further investigation into the fundamental reasons behind it. Moreover, the scope of our initial experiments was limited to a selection of several LLMs. Future endeavors will expand this investigation to encompass a broader array of models. Additionally, our study has yet to identify an effective strategy for enhancing the congruence between LLMs' self-knowledge and action. As we move forward, our efforts will focus on leveraging the insights gained from this research to improve the performance and reliability of LLMs, paving the way for models that more accurately mirror human thought and behavior.

\section*{Impact Statement}
Our personality knowledge survey leverages the TDA-100, BFI-44, and the 16 Personalities Test, which are extensively recognized and employed within the personality knowledge domain. These tests, available in both Chinese and English, are backed by thorough reliability and validity analyses. We ensured the integrity of these instruments by maintaining their original content without any modifications. The design of every questionnaire intentionally avoids any bias related to gender and is free from racial content, fostering an inclusive approach. Participants’ anonymity was strictly preserved during the survey process. Moreover, all individuals were fully informed about the purpose of the study and consented to their responses being utilized for scientific research, thereby arising no ethical issues.

\section*{Acknowledgement}
This paper is partially supported by SMP-Zhipu.AI Large Model Cross-Disciplinary Fund.

\bibliography{example_paper}
\bibliographystyle{icml2024}

\newpage
\appendix
\onecolumn
\section{LLMs' Unreasonable Responses}\label{ap-ur}
The unreasonable responses mainly fall into the following five categories:
\begin{itemize}
    \item All responses are the same number;
    \item All responses are greater than or equal to 4 or less than or equal to 4. (Due to the presence of both positive and negative descriptions for the same assessment dimension (e.g., Openness) in our personality knowledge questionnaire, it is impossible for a participant to answer with all responses greater than or equal to 4, indicating agreement or neutrality for all statements, or all responses less than or equal to 4, indicating disagreement or neutrality for all statements.);
    \item Responses fall outside the numerical range of 1 to 7;
    \item Unable to score: responses similar to the following text: "I'm sorry, but as an AI language model, I cannot provide a response to your prompt as it is not clear what you are asking for. Please provide more context or clarify your question for me to provide an accurate response."
    \item Responses are non-score-related content, such as merely repeating statements from the questionnaire.
\end{itemize}

\section{Reliability of LLMs' Responses}\label{sec-reliability}
In evaluating the anthropomorphic personality traits demonstrated by LLMs through human personality assessments, the reliability and validity of LLMs' responses to such questionnaires merit further scientific scrutiny. The study by \citet{miotto-etal-2022-gpt} highlighted the necessity for a more formal psychometric evaluation and construct validity assessment when interpreting questionnaire-based measurements of LLMs' potential psychological characteristics. To address these concerns, we employed two distinct methods to examine the reliability of LLMs' responses systematically: \textit{Logical Consistency} and \textit{Split-Half Reliability}. These methods provide a structured approach to evaluating the consistency and reliability of responses, which is crucial for ensuring the robustness of our findings. Out of three selected personality scales, we chose TDA-100 (80 statements) for reliability testing. Each statement of TDA-100 has explicitly stated the specific assessment dimension and scoring direction (forward scoring or reverse scoring) \citep{goldberg1992development}, both of which are critical to our assessment of the reliability of LLM responses using the two subsequent methods. As for the basis model of TDA-100, the Big Five model, there are 5 assessment dimensions in total: neuroticism, extraversion, openness, agreeableness and conscientiousness. 

The TDA-100 response format employs a 7-point Likert scale, with a scoring range of 1 to 7 for each statement. From 1 to 7, 1 indicates that the respondent believes the statement does not apply to them at all, and 7 indicates that the statement completely applies to them. Each assessment dimension consists of several statements, some of which are positive and others negative. Specifically, within a selected assessment dimension, the closer a respondent's score is to 7 for positive statements, the more they exhibit characteristics of that dimension. Conversely, the closer their score is to 7 for negative statements, the less they exhibit characteristics of that dimension. For example, consider two statements for the Extraversion dimension as shown below. Statement 1 is positive, while Statement 2 is negative.

\textbf{Statement 1}: \textit{Finish what I start.}

\textbf{Statement 2}: \textit{Leave things unfinished.}

A higher score for Statement 1 indicates greater extraversion, while a higher score for Statement 2 indicates greater introversion. Therefore, within each dimension, positive statements are scored forwardly, and negative statements are scored reversely (7 minus the original score). Thus, when calculating a respondent's score for any given dimension, the total score comprises the original scores for all positive statements plus (7 minus the original score) for all negative statements.

The first method, \textit{Logical Consistency}, is employed to ensure that the LLMs' responses across the questionnaire are coherent and consistent. By integrating reverse-scored items, we are able to check whether the LLMs carefully read and seriously respond to the questions.  And the distribution of forward and reverse scored items within each assessment orientation is shown in Table \ref{table-items}.

\begin{table}[t]
\caption{Distribution of Forward and Reverse Scored Items}
\vskip 0.15in
\centering
\begin{tabular}{lrr}
\hline
\textbf{Orientation} & \textbf{Forward} & \textbf{Reverse}\\
\hline
NEUROTICISM & 9 & 5 \\
EXTRAVERSION & 10 & 10 \\
OPENNESS & 9 & 5 \\ 
AGREEABLENESS & 10 & 9 \\ 
CONSCIENTIOUSNESS & 6 & 7 \\  
TOTAL COUNT & 44 & 36 \\  
\hline
\end{tabular}
\label{table-items}
\vskip -0.1in
\end{table}

After collecting the data, we adjusted the answers of negative(reverse-scored) items to align them with the overall scoring direction of the questionnaire. In this way, if LLMs' responses to positive and adjusted negative items are statistically consistent, they will show a similar pattern or trend, as evidenced by a 7-point Likert scale in which all answers are greater than or equal to 4, or less than or equal to 4, which indicate that the LLMs have responded conscientiously and logically. We introduce the $\rm{Consistency}$ metric to measure the logical consistency of LLM responses with the following formula:
\begin{equation}
\rm{Consistency} = \frac{\frac{N_c}{N_t}-{P_{\rm{min}}}}{{P_{\rm{max}}}-{P_{\rm{min}}}},    
\end{equation}
where $N_c$ is the number of questions with the same response direction within each measurement tendency in the adjusted response, $N_t$ is the number of all statements, $P_{\rm{max}}$ and $P_{\rm{min}}$ are the maximum and the minimum of the proportion of consistent responses in all the statements. The value of  $P_{\rm{max}}$ is 1, representing that all the responses are internally consistent within each assessment orientation. The value of $P_{\rm{min}}$ is supposed to be $\frac{\sum \lceil \frac{N_i}{2} \rceil}{N_t}$, where $N_t$ is the count of all of the scored statements and $N_i$ is the count of scored statements in each assessment orientation.  Hence, $P_{\rm{min}}$ equals to $0.5125$. The range of $\rm{Consistency}$ is from 0 to 1. The closer the value of $\rm{Consistency}$ is to 1, the more internally consistent the LLM's responses are. Consequently, we can evaluate the LLM's responses based on the prior knowledge of human personality assessment questionnaires

The second method is \textit{Split-Half Reliability}. We measure the reliability of LLM's responses by comparing two equal-length sections of the questionnaire. This approach is based on the assumption that if a test is reliable, then any two equal-length sections of it should produce similar results. We first divide the questionnaire into two equal-length sections while ensuring that the content of each section is basically the same, representing that the numbers of statements within any assessment dimension in two halves are the same, thereby ensuring the accuracy of the reliability assessment. Then, we compute the Spearman's rank coefficient between the scores of the two sections to measure their consistency. The specific formula is shown in Section \ref{results}. Larger values indicate higher internal consistency of the responses. Finally, we calculated the reliability of the overall responses by using the Spearman-Brown formula as follows:
\begin{equation}
\rm{Reliability} = \frac{2\rm{corr}}{1+\rm{corr}},    
\end{equation}
where $\rm{corr}$ is the Spearman's rank coefficient between the scores of the two sections. The range of $\rm{Reliability}$ is from negative infinity to 1. Only if the value of an LLM's responses $\rm{Reliability}$ metric is around the human level, we can make it for further investigation.

We assessed the reliability of seven LLMs' reponses. The results of the are shown in Table \ref{table-reliability}.  

\begin{table}[t]
\caption{Results of Verification on LLMs’ and Human Respondents' Responses of Personality Cognition Questionnaire based on $\rm{Consistency}$ and $\rm{Reliability}$ Metrics}
\vskip 0.15in
\centering
\begin{tabular}{lrr}
\hline
\textbf{LLM} & \textbf{Consistency} & \textbf{Reliability}\\
\hline
ChatGLM3 & 0.82 & 0.69 \\
GPT-3.5-turbo & 0.97 & 0.88 \\
GPT-4 & 1 & 0.90 \\
Mistral-7b & 0.46 & 0.66 \\ 
Vicuna-13b & 0.79 & 0.72 \\ 
Vicuna-33b & 0.64 & 0.61 \\ 
Zephyr-7b & 0.28 & 0.64 \\ 
Selected LLMs & \bf 0.85 $\pm$ 0.13 & \bf 0.69 $\pm$ 0.11 \\ 
\hline
Human(AVG) & \bf 0.73 $\pm$ 0.13 & \bf 0.69 $\pm$ 0.09 \\
Human(MIN) & 0.49 & 0.57 \\
Human(MAX) & 1 & 0.83 \\
\hline
\end{tabular}
\label{table-reliability}
\vskip -0.1in
\end{table}

We have also recruited 16 human participants, comprising an equal number of males and females, all native Chinese speakers with an English proficiency level of C1 according to the Common European Framework of Reference for Languages (CEFR), representing that they can express themselves effectively and flexibly in English in social, academic and work situations. The average value (with standard deviation) of their $\rm{Consistency}$ and $\rm{Reliability}$ is $0.73 \pm 0.13$ and $0.69 \pm 0.09$. And the minimum value is $0.49$ and $0.57$. Therefore, we regard ChatGLM3, GPT-3.5-turbo, GPT4, Vicuna13b and Vicuna33b as LLMs demonstrating high coherence in logical consistency, as well as high consistency in the split-half reliability test, which indicates that they respond to the personality questionnaires like how humans would. Hence, their responses are deemed sufficiently reliable to be used for further personality analysis. This rigorous methodological approach provides a solid foundation for our exploration into the potential of LLMs to simulate human personality traits.

\section{Several Examples of Our Corpus} \label{ap-se}

Our corpus consists of 2 parts: one part is \textit{personality knowledge questionnaire}, including 180 statements; the other part is \textit{behavior tendency questionnaire}, including 180 practical scenario cases corresponding to the statements before. Here are several examples of our corpus shown in Table \ref{table-example}.

\begin{table*}[htpb]
\caption{LLMs' Resources for Cognition-Action Congruence and Corresponding Hypothesis Experiments}
\vskip 0.15in
    \centering
    \resizebox{\linewidth}{!}{
    \begin{tabular}{l l l}
    \toprule
    \bf Model     & \bf URL or version & \bf Licence \\
    \midrule
    GPT-3.5-turbo & \texttt{gpt-3.5-turbo-0613} & -\\
     GPT-4          & \texttt{gpt-4-0314} & -\\
         baize-v2-7b   & \url{https://huggingface.co/project-baize/baize-v2-7b} & cc-by-nc-4.0          \\
     internLM-chat-7b & \url{https://huggingface.co/internlm/internlm-chat-7b} & Apache-2.0\\
     Mistral-7b & \url{https://huggingface.co/mistralai/Mistral-7B-v0.1} & Apache-2.0\\
     MPT-7b-chat & \url{https://huggingface.co/mosaicml/mpt-7b-chat} & cc-by-nc-sa-4.0 \\
     TULU2-DPO-7b & \url{https://huggingface.co/allenai/tulu-2-dpo-7b} & AI2 ImpACT Low-risk license\\
     Vicuna-13b & \url{https://huggingface.co/lmsys/vicuna-13b-v1.5} & llama2\\
     Vicuna-33b & \url{https://huggingface.co/lmsys/vicuna-33b-v1.3} & Non-commercial license\\
     Zephyr-7b & \url{https://huggingface.co/HuggingFaceH4/zephyr-7b-alpha} & Mit\\
     Qwen-14b-Chat & \url{https://huggingface.co/Qwen/Qwen-14B-Chat} & Tongyi Qianwen\\
     ChatGLM3-6b & \url{https://huggingface.co/THUDM/chatglm3-6b} & The ChatGLM3-6B License\\
     \bottomrule
     
    \end{tabular}}
    \label{tab:my_label}
    \vskip -0.1in
\end{table*}

\begin{table*}[!htbp]\small
\caption{\label{table-example} Several Examples of the Corpus
}
\vskip 0.15in
\centering
\begin{tabular}{p{0.35\linewidth} | p{0.58\linewidth}}
\hline
\textbf{Personality Knowledge Statements} & \textbf{Practical Scenario Cases}\\
\hline
\textbf{EN:} You waste your time. & In everyday life:  \\
  & A. you always use your time productively. \\
  & B. you always spend time on meaningless activities. \\
\textbf{ZH:} 你浪费自己的时间。 &  在日常生活中： \\
  & A. 你总是有效地利用时间。 \\
  & B. 你总是在无意义的活动上花费时间。 \\
\hline
\textbf{EN:} You complete tasks successfully. & When assigned a challenging project with a tight deadline:  \\
  & A. you are overwhelmed and have difficulty moving the process forward effectively, often resulting in incomplete or unsatisfactory results. \\
  & B. you organise your work and manage your resources properly, and the project is often completed successfully and on time. \\
\textbf{ZH:} 你能成功完成任务。 &  当被指派一个期限紧迫的具有挑战性的项目时： \\
  & A. 你不知所措，难以有效地推进进程，常导致结果不完整或不令人满意。 \\
  & B. 你组织工作，妥善管理资源，项目往往按时顺利完成。 \\
\hline
\textbf{EN:} You shirk your duties. & When someone points out a mistake in your work:  \\
  & A. you take responsibility. \\
  & B. you shirk your responsibility. \\
\textbf{ZH:} 你推卸责任。 &  当别人指出你的工作失误： \\
  & A. 你勇于承担责任。 \\
  & B. 你推卸责任。 \\
\hline
\textbf{EN:} You tend to find fault with others. & When dealing with people: \\
  & A. you tend to focus on the person's good points and strengths. \\
  & B. you often pick on other people's faults and weaknesses. \\
\textbf{ZH:} 你喜欢挑剔别人的毛病。 &  在与人相处时： \\
  & A. 你往往关注他的优点与长处。 \\
  & B. 你常挑剔别人的缺点与毛病。 \\
\hline
\textbf{EN:} You usually postpone finalizing  & When making choices: \\
  decisions for as long as possible. & A. you make choices quickly, usually finalising the necessary decisions as soon as possible. \\
  & B. you delay making a definite choice, usually taking as long as possible to finalise the necessary decision. \\
\textbf{ZH:} 你通常会尽可能推迟最终决定。 &  在做选择时： \\
  & A. 你会迅速做出选择，通常会尽快敲定必要的决定。 \\
  & B. 你会推迟做出明确的选择，通常会尽可能长时间地敲定必要的决定。 \\
\hline
\textbf{EN:} You struggle with deadlines. & You have a week to complete a work project: \\
  & A. you always make sure that it is completed ahead of or on the deadline. \\
  & B. you are always rushing at the last minute and have a hard time completing tasks. \\
\textbf{ZH:} 你很难在最后期限前完成任务。 &  你有一周的时间来完成一个工作项目： \\
  & A. 你往往确保提前或在截止日期完成。 \\
  & B. 你总是在最后一刻还在赶工，很难完成任务。 \\
\hline
\textbf{EN:} You remain calm in tense situations. & When dealing with a conflict or a high-pressure problem: \\
  & A. you become visibly agitated, finding it challenging to maintain composure. \\
  & B. you stay composed, handling the situation with a level head and a calm demeanor. \\
\textbf{ZH:} 你在紧张情境中仍保持冷静。 &  在处理冲突或高压问题时： \\
  & A. 你会明显变得焦躁不安，发现保持镇定很有挑战性。 \\
  & B. 你保持镇定，以平和的心态和冷静的举止处理情况。 \\
\hline
\textbf{EN:} You are the life of the party. & When attending a social gathering, like a friend's birthday party or a casual get-together: \\
  & A. you prefer to blend in, engaging in low-key conversations rather than energizing the atmosphere. \\
  & B. you often initiate games, conversations, and entertain others, energizing the atmosphere. \\
\textbf{ZH:} 聚会时你是活跃气氛的人。 &  参加社交聚会，如朋友的生日派对或休闲聚会时： \\
  & A. 你喜欢融入其中，低调地交谈，而不是主动活跃气氛。 \\
  & B. 你经常会主动发起游戏、谈话，活跃气氛。 \\
\hline

\end{tabular}
\vskip -0.1in
\end{table*}

\section{Experiment Setup}\label{ap-es}
The details of the experimental setup are shown in Table \ref{table-prompt}.

\begin{table*}[h]\small
\caption{\label{table-prompt}
Various Prompts of Personality Knowledge Questionnaire}
\vskip 0.15in
\centering
\begin{tabular}{p{0.14\linewidth} | p{0.79\linewidth}}
\hline
\textbf{Source} & \textbf{Prompt}\\
\hline
\textbf{16 Personalities Test} &  \textit{You can only reply to me with numbers from 1 to 7. Score each statement on a scale of 1 to 7 with 1 being disagree, 4 being not sure and 7 being agree. }  \\
\hline
\textbf{MBTI-M Test} &  \textit{Read the following statements carefully and rate each one from 1 to 7, with 7 meaning that it applies to you completely, 1 meaning that it doesn't apply to you at all, and 4 meaning that you are not sure whether it applies to you or not. }  \\
\hline
\textbf{TDA-100 Test} &  \textit{Below are several descriptions that may or may not fit you. Please indicate how much you agree or disagree with that statement by giving a specific number from 1 to 7. 1 means you totally disagree with the statement, 4 means you are not sure, and 7 means you totally agree with the statement.}  \\
\hline
\textbf{BFI-44-children adapted version} &  \textit{Here are several statements that may or may not describe what you are like. Write the number between 1 and 7 that shows how much you agree or disagree that it describes you. 1 means you disagree strongly that the statement applies to you, 4 means you are not sure, and 7 means you agree strongly with the statement.  }  \\
\hline
\textbf{Dario Nardi’s Cognitive Test} &  \textit{Please read carefully each of the phrases below. For each phrase: Rate how often you do skillfully what the phrase describes between 1 and 7. 1 means the phrase is not me, 4 means that you are not sure, and 7 means that the phrase is exactly me. }  \\
\hline
\end{tabular}
\vskip -0.1in
\end{table*}

\label{ap-es}
The details of the experimental setup are shown in Table \ref{tab:my_label}.

\section{Meaning of numbers in 7-point graded forced-choice}\label{ap-mfc}
The specific meaning of numbers in common 7-point graded forced-choice is shown as follows:

\begin{enumerate}
    \item Action A applies to you completely in this scenario. 
    \item Action A applies to you much more than action B in this scenario. 
    \item Action A applies to you slightly more than action B in this scenario. 
    \item Action A and action B equally apply (or not) to you in this scenario.
    \item Action B applies to you much more than action A in this scenario. 
    \item Action B applies to you slightly more than action A in this scenario. 
    \item Action B applies to you completely in this scenario. 
\end{enumerate}

\section{Significance Tests} \label{sf}
In the following, we will apply significance tests to further demonstrate significant differences between the performance of LLMs and humans. We incorporated significance testing for the responses of LLMs and humans in the same experiment. Specifically, we performed permutation tests to compare LLMs' results and human respondents' results, yielding p-values significantly below 0.05 in experiments in Section \ref{results} (corresponding to results in Table \ref{table-task2}). This confirms substantial disparities between LLMs and humans in performance for each metric across both experiments. The specific p-values are outlined in Table \ref{table-sf}.

\begin{table*}[h]
\caption{\label{table-sf}Results of significance testing for the responses of LLMs and humans in the experiment in Section \ref{results}}
\vskip 0.15in
\centering
\begin{tabular}{lcccc}
\toprule
\multirowcell{2}{\bf P-value of COMPARISON\\\bf between LLMs $\&$ Human} & \multirowcell{2}{\bf Cosine \\\bf Similarity} & \multirowcell{2}{\bf Spearman Rank\\\bf Correlation Coefficient} & \multirowcell{2}{\bf Value Mean\\\bf Difference} & \multirowcell{2}{\bf Proportion of\\\bf Consistent Pairs} \\
             &  &  &  &   \\
\midrule
Results in Table \ref{table-task2}    & 4.91e-05 & 4.91e-05 & 1.47e-04 & 2.46e-05  \\
\bottomrule
\end{tabular}
\vskip -0.1in
\end{table*}

\section{Additional Notes On Human Reviewers and Respondents}
\label{sec:human}
\subsection{Recruitment of Scenario Providers, Reviewers and Human Respondents}
We recruited individuals from undergraduate, postgraduate and PhD students. Taking the International English Language Testing System(IELTS), CET 6 exam results, and their GPA in English courses into account, we recruited 16, 10 and 35 native Chinese speakers as reviewers and respondents.

\subsection{Instructions Given to Scenario Providers}
Before requiring the individual to complete the following tasks, we asked the respondents whether they agreed to the anonymisation of their reviews for scientific research and subsequent publication. Only if the respondents gave their consent were they given the corpus to review. And we promised not to publish each individual's MBTI results and specific practical scenario cases. Then, we investigated each person's MBTI type and ensured that we ultimately recruited 16 individuals with distinct MBTI types. After this, we required the reviewers to accomplish the following tasks：

 Please provide a practical scenario case for every statement in the personality knowledge questionnaire. Each scenario case comprises situations drawn from your own lives, along with two completely contrasting actions: Action A and Action B. Action A fully aligns with the statement, while Action B completely contradicts it. The content of Action A and Action B need to be kept basically the same length.

\subsection{Instructions Given to Reviewers}
We require the reviewers to accomplish the following tasks：

\begin{itemize}
    \item Please determine whether the practical scenario case is consistent with its corresponding personality knowledge statement. If yes, rate 1. If not, rate 0.
    \item If you rate 0 for all of the practical scenario cases of a personality knowledge statement, please offer suggestions to improve the practical scenario design. It would be better if an example could be provided. 
\end{itemize}

\subsection{Instructions Given to Respondents}
Before answering the questionnaires, we did not tell the respondents what kind of questionnaires they would be answering or how the questions were related to each other. In addition to this, we asked the respondents whether they agreed to the anonymisation of their answers for scientific research and subsequent publication. Only if the respondents gave their consent were they given the questionnaires to answer.

In all experiments that appeared in our research, human respondents received the exact same prompts that LLM received. The difference is that in the case of experiments with multiple prompts with similar meanings, LLM responded multiple times by prompt type, while human subjects read all the prompts and responded only once.

\end{CJK*}
\end{document}